\documentclass[conference,a4paper]{IEEEtran}

\usepackage{amsmath,epsfig}
\usepackage{multicol,multirow,color}
\usepackage{booktabs}
\usepackage{colortbl}
\usepackage{tabularx}
\usepackage{subcaption}
\usepackage{hyperref}
\usepackage[font=small,labelfont=bf,tableposition=top]{caption}
\usepackage{breakurl}

\begin{document}
\title{Towards Universal GAN Image Detection}

\author{
\IEEEauthorblockN{Davide Cozzolino, Diego Gragnaniello, Giovanni Poggi, Luisa Verdoliva}
\IEEEauthorblockA{University Federico II of Naples, Italy\\ Email: name.surname@unina.it
}
}

%\maketitle
\twocolumn[{%
	\renewcommand\twocolumn[1][]{#1}%
	\maketitle
	\begin{center}
	    \includegraphics[page=1, width=1\linewidth, trim = 25 280 140 0, clip]{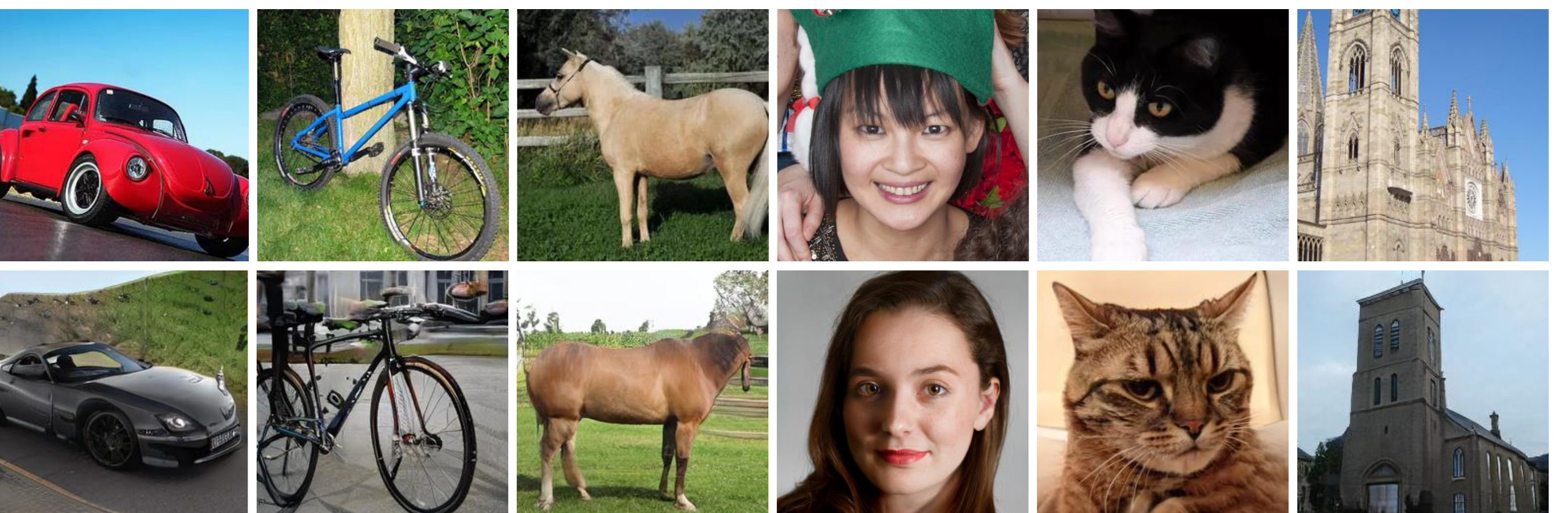}
		\captionof{figure}{Pristine (Top) and synthetic (Bottom) images generated using ProGAN \cite{karras2018progressive} and StyleGAN2 \cite{karras2020analyzing} architectures.}
		\label{fig:teaser}
	\end{center}
}]

\begin{abstract}
The ever higher quality and  wide diffusion of fake images have spawn a quest for reliable forensic tools. Many GAN image detectors have been proposed, recently. 
In real world scenarios, however, most of them show limited robustness and generalization ability. Moreover, they often rely on side information not available at test time, that is, they are not universal.
We investigate these problems and propose a new GAN image detector based on a limited sub-sampling architecture and a suitable contrastive learning paradigm. Experiments carried out in challenging conditions prove the proposed method to be a first step towards universal GAN image detection, ensuring also good robustness to common image impairments, and good generalization to unseen architectures.

\end{abstract}

\IEEEpeerreviewmaketitle

\section{Introduction}
Image synthesis based on GAN technology has apparently reached full maturity.
Two recent studies \cite{Nightingale2021synthetic, Lago2021more} have independently proved that humans cannot reliably tell apart images generated by advanced GAN technologies, such as StyleGAN2 \cite{karras2020analyzing}, from pristine images.
The average accuracy turned out to be around 50\% (coin tossing) for untrained observers,
increasing to just 60\% for trained observers with unlimited analysis time \cite{Nightingale2021synthetic}.
This raises legitimate concerns about the use of GAN images for the most different purposes. 
Besides plain jokes, such as the fictional profile of a U.S. Congress candidate built on Twitter \cite{Faketwitter} by a seventeen years old kid,
there is also the growing spread of government propaganda over Facebook \cite{gov_propaganda_fb}.
In addition,
while early efforts were almost exclusively on faces, recent GANs can realistically generate from scratch many other categories, as shown in Fig.1 where real and high-quality synthetic images are shown.

Yet, while GAN images can evade the scrutiny of human observers,
their generation process leaves distinctive traces \cite{Marra2019dogan, Yu2019attributing} which allow detection by means of dedicated forensic tools.
In the literature, there has been an intense research effort for reliable GAN image detectors \cite{Verdoliva2020media}, 
and many of them have shown impressively good results when suitably trained.

However, one of the main challenges in GAN image detection is generalization.
In fact, if a detector is tested on the very same type of images seen in the training phase it will hardly fail.
Unfortunately, this is not what happens in a realistic scenario.
To operate successfully in the wild, a detector should {\it i)} be robust to image impairments, {\it ii)} work universally well across sources and {\it iii)} generalize well to new sources.
Robustness to image impairments is essential, since most social networks resize and compress images to satisfy their internal constraints.
These non-malicious operations destroy precious evidence, with a possible detrimental effect on detection performance.
Universality is also a fundamental requirement, since the detector ignores the image source (which GAN architecture was used to generate it) and cannot use this information.
Finally, the image under test may have been generated by a totally new architecture, and the detector should work well also in this condition.

The issue is touched upon in \cite{Wang2020} where
it has been also shown that a neural network trained to detect images generated by a given GAN generalizes well to image generated by other GANs, provided suitable augmentation is used in training.
Experiments prove the proposed detector to ensure a large AUC (area under the ROC curve) figure on GAN images never seen in training.
The AUC, however, measures the {\it potential} of the method rather than its actual performance.
A large AUC ensures that scores for real and fake images are reasonably separable, and reliable detection is possible given a good decision threshold.
However, the decision threshold itself is unknown and, more important, it varies largely from GAN to GAN.
In \cite{Wang2020} it is observed that a good decision threshold can be estimated even from just two images, one real and one fake.
However, in real-world testing conditions, the detector does not know which GAN possibly generated the image under test and hence which threshold to use.

Here, we propose a contrastive learning based architecture that ensures a more stable performance across different sources.
Experiments carried out in real-world conditions, with performance metrics less forgiving than the AUC, show the proposed detector to outperform all reference methods.

\section{Related work}
In recent years, there has been intense research on forensic detectors that distinguish between real and synthetic content.
Leveraging the opportunity to generate large datasets of GAN images, the most effective techniques work in a supervised setting and achieve excellent results both working in the spatial \cite{dang2020detection, chai2020makes} and in the frequency domain \cite{Zhang2019, frank2020leveraging}.
These detectors provide an excellent performance in ideal conditions, that is, with high-quality images that come from a GAN architecture included in the training set.
Often, however, the performance degrades sharply in more challenging real-world scenarios,
with images that are resized and compressed, and may be generated by GAN architectures that were not included in training. 
To improve generalization, \cite{Cozzolino2018} proposes a few-shot learning method, while an approach based on incremental learning is considered in \cite{Marra2019incremental}.
However, in both cases the improved generalization is obtained by exploiting a few example images from the new GAN architectures,
pieces of information hardly available in a real-world scenario.

The most successful tool to gain both generalization and improved robustness is augmentation.
In \cite{xuan2019generalization} the idea is to carry out augmentation by Gaussian blurring, so as to force the detector to learn more general features while discarding noise-like patterns that impair the training.
A similar approach is followed in \cite{Wang2020} where a standard pre-trained model, ResNet50, is further trained with a strong augmentation based on compression and blurring. 
Experiments show that, even when the detector is trained on images from a single GAN architecture, the learned features generalize well to unseen architectures, datasets, and training methods. 
This approach is further analyzed in \cite{gragnaniello2021gan} where several variations are considered.
In particular, inspired by recent work in steganalysis \cite{yousfi2020imagenet} and image forensics \cite{marra2020full}, an architecture where down-sampling operations are removed from the first layer of the network is proposed.
The promising results confirm the need to preserve the subtle traces hidden in high-frequency image components for forensic applications.

\begin{figure}[t!]
    \centering
    \includegraphics[width=1.0\linewidth,page=2, trim = 0 340 215 0, clip]{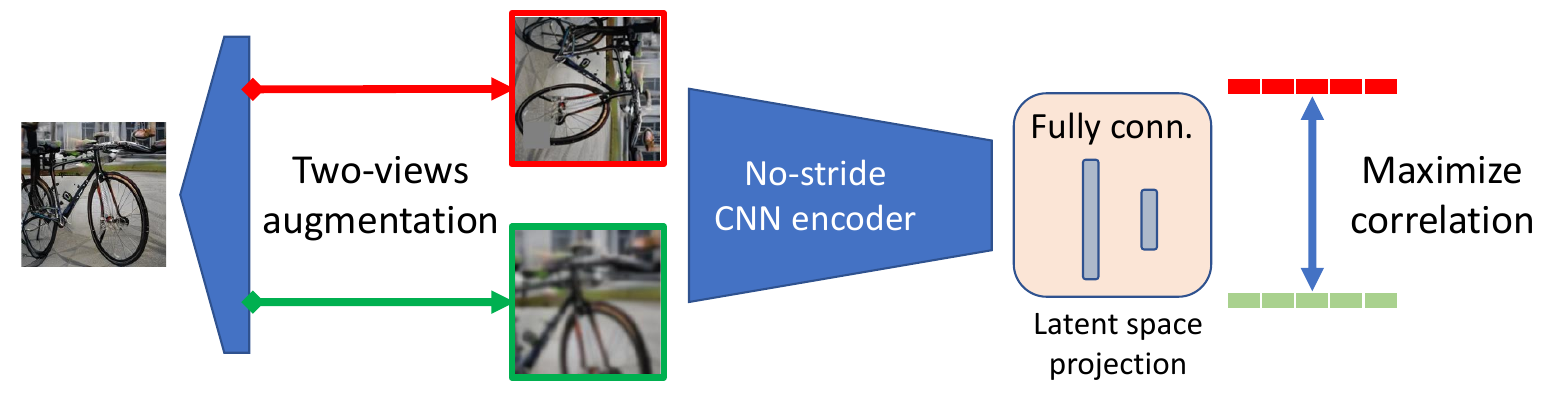}
    \caption{Scheme of our contrastive learning based approach.}
    \label{fig:contrastive}
\end{figure}

\section{Proposed method}

Here we propose a new GAN image detector aimed at ensuring good generalization and robustness also in real-world scenarios.
We build upon the lessons learnt in \cite{Wang2020} and in \cite{gragnaniello2021gan} and accordingly try to
{\it  i)} avoid early subsampling of input data, and
{\it ii)} use intense augmentation of the training data.

In terms of architecture, we use the same backbone adopted in \cite{Wang2020}, a simple ResNet50 pretrained on ImageNet but, as proposed in \cite{gragnaniello2021gan}, we do not perform subsampling in the first layer.
As for augmentation, we turn to a contrastive learning paradigm \cite{Chen2020simple,Khosla2020supervised}, suitably adapted to our peculiar needs.
More specifically, we adopt the two-step training procedure proposed in \cite{Chen2020simple} for visual representation learning.
We begin by removing the last fully-connected layers from our ImageNet pre-trained backbone.
The feature vector computed by the network after the global average pooling is then projected in a new latent space for a first self-supervised training phase.
This crucial step aims at extracting a representation of the input image that is as invariant as possible to various forms of distortion and impairment.
To this end, the network is fed with multiple augmented versions of the same image, and forced to generate similar latent vectors for them by minimizing a suitable contrastive loss.
In particular, we use the Normalized Temperature-scaled Cross Entropy (NT-Xent) loss proposed originally in \cite{Chen2020simple}.
Let $u$ and $u^+$ be the latent vectors extracted from two different augmented versions of the original image I, and let $v$ indicate either the latent vector $u^+$ or a generic latent vector extracted from another image, unrelated to I, then
\begin{equation}
    \mbox{Loss} =  - sim(u,u^+) / \tau + \log \sum_{v \ne u} \exp \left(  sim(u,v) /\tau   \right)
\end{equation}
Here, $sim(u,v) = u^T v / ( \left\|u\right\| \left\|v\right\|)$ is the cosine similarity (the dot product between normalized $u$ and $v$),
the summation on $v$ includes the compact feature vectors extracted by all images of the batch except $u$,
and $\tau$ is a temperature parameter, set to 0.07 in our experiments.
By minimizing the NT-Xent loss, the network increases the correlation between $u$ and $u^+$ as opposed to the correlation between $u$ and the other latent vectors of the batch.

This approach fits very well our needs, and we use a large set of augmentations which reproduce most of the post-generation processing applied to synthetic images, for example when they are posted on a social network. 
Beyond JPEG compression and Gaussian blurring, already adopted in \cite{Wang2020}, we also randomly vary the brightness, contrast, saturation, and hue, as well as random gray-scale conversion. Also, images are randomly distorted adding Gaussian noise or removing patches via CutOut. Then, random patches of dimension $96\times96$ are extracted during training. 
Instead, at test time we perform inference on full-resolution images thanks to the global average pooling.
It is worth emphasizing that both at training and at test time no resizing is carried out in order to preserve information and avoid destroying important traces. 

Subsequently,
we add the fully-connected layers and fine-tune the network in a supervised manner for our final task, that is synthetic image detection.
Note that, unlike in \cite{Chen2020simple}, we fine-tune the whole network and not only the last layers.
In fact, we are interested in all the low- and mid-level traces left by the GANs in the generation process, and not only to high-level visual representation. 

As for the implementation details,
to project the image representation in the latent space for the self-supervised learning, we use two fully-connected layers of size 2048 and 128, interleaved by rectified linear unit activation.
During the self-supervised training, the network parameters are updated via stochastic gradient descent with an initial learning rate of 1e-4, then reduced by a factor of 10 until 1e-6 if the validation loss does not decrease after 5 consecutive epochs, and a batch size of 64 images (2 different views obtained from 32 original images). 
Finally, during the supervised fine-tuning, the Adam \cite{Kingma2015Adam} optimizer is adopted with standard parameters, an initial learning rate of 1e-5 and the same learning rate scheduling and batch size of the self-supervised training.

\newcommand{\ru}{\rule{0mm}{3mm}}
\begin{table*}[t!]
    \centering
    \begin{tabular}{|cl|ccccccc|}
    \cline{3-9}
\multicolumn{2}{c|}{\ru ACC / AUC} & Spec & FFD & PatchForensics & CNN-DCT & Wang2020 & no-down & proposed \\ \cline{1-9}
\ru  \multirow{8}{*}{ \rotatebox{90}{Low R.}}
    &        ProGAN    & ~78.3 /0.987 & ~85.1 /0.977 & ~66.1 /0.982 &  ~60.0 /0.664 & ~99.3 /1.000 & ~94.7 /1.000 & ~99.4 /1.000 \\ 
\ru &      StyleGAN    & ~63.6 /0.704 & ~70.2 /0.761 & ~59.2 /0.959 &  ~53.2 /0.478 & ~75.9 /0.954 & ~93.7 /0.988 & ~98.2 /0.999 \\
\ru &     StyleGAN2    & ~53.8 /0.557 & ~64.6 /0.726 & ~53.0 /0.933 &  ~29.8 /0.270 & ~71.5 /0.946 & ~92.2 /0.980 & ~89.0 /0.990 \\
\ru &        BigGAN    & ~77.4 /0.929 & ~57.4 /0.640 & ~52.0 /0.867 &  ~60.0 /0.697 & ~59.2 /0.900 & ~93.5 /0.985 & ~95.3 /0.996 \\
\ru &      CycleGAN    & ~78.2 /0.961 & ~67.8 /0.733 & ~63.7 /0.833 &  ~60.1 /0.700 & ~77.4 /0.968 & ~90.3 /0.969 & ~94.0 /0.994 \\
\ru &       StarGAN    & ~77.0 /0.819 & ~85.3 /0.987 & ~98.5 /1.000 &  ~60.1 /0.703 & ~84.3 /0.978 & ~94.5 /0.992 & ~98.6 /0.999 \\
\ru &        RelGAN    & ~78.3 /0.992 & ~85.3 /0.971 & ~99.3 /1.000 &  ~60.1 /0.703 & ~63.6 /0.914 & ~92.8 /0.977 & ~96.4 /0.997 \\
\ru &        GauGAN    & ~76.8 /0.890 & ~64.5 /0.727 & ~50.0 /0.751 &  ~60.1 /0.703 & ~82.5 /0.975 & ~93.6 /0.988 & ~94.5 /0.996 \\ \hline \hline
\ru \multirow{3}{*}{ \rotatebox{90}{High R.~}}
    & ProGAN           & ~83.3 /0.936 & ~69.2 /0.962 & ~97.9 /1.000 &  ~66.9 /0.668 & ~99.7 /1.000 & ~97.1 /1.000 & ~95.3 /1.000 \\
\ru & StyleGAN         & ~87.3 /0.989 & ~68.0 /0.877 & ~50.6 /0.954 &  ~59.7 /0.647 & ~91.0 /0.993 & ~96.8 /0.997 & ~95.3 /1.000 \\ 
\ru & StyleGAN2        & ~87.8 /0.999 & ~68.8 /0.867 & ~50.0 /0.912 &  ~55.5 /0.540 & ~73.2 /0.960 & ~96.9 /0.998 & ~95.1 /0.997 \\ \hline \hline
\ru & avg.             & ~76.5 /0.888 & ~71.5 /0.839 & ~67.3 /0.927 &  ~56.7 /0.616 & ~79.8 /0.963 & ~94.2 /0.989 & ~{\bf 95.6 /0.997} \\ 
\hline  
    \end{tabular}
    \vspace{2mm}
    \caption{Accuracy and AUC for all the methods that have been trained only on ProGAN dataset.}
    \label{tab:AUC-Acc}
\end{table*}

\begin{table*}[t!]
    \centering
    \begin{tabular}{|cl|ccccccc|}
    \cline{3-9}
\multicolumn{2}{c|}{\ru Pd@1\% / Pd@10\%} & Spec & FFD &  PatchForensics & CNN-DCT & Wang2020 & no-down & proposed \\ \cline{1-9}
\ru \multirow{8}{*}{ \rotatebox{90}{Low R.}}
    &        ProGAN    & ~79.6 / 98.0 & ~61.3 / 94.2 & ~66.0 / 96.8 &  ~ 1.5 / 15.0 & 100.0 /100.0 & 100.0 /100.0 & 100.0 /100.0 \\ 
\ru &      StyleGAN    & ~ 7.9 / 29.8 & ~23.8 / 56.0 & ~43.6 / 91.1 &  ~ 0.8 /  8.4 & ~49.4 / 85.0 & ~76.9 / 97.8 & ~97.3 /100.0 \\
\ru &     StyleGAN2    & ~ 4.1 / 17.8 & ~ 9.6 / 36.1 & ~26.7 / 82.3 &  ~ 0.4 /  3.8 & ~40.6 / 82.0 & ~68.4 / 94.7 & ~77.6 / 98.5 \\
\ru &        BigGAN    & ~36.9 / 77.3 & ~ 2.2 / 15.8 & ~12.4 / 62.7 &  ~ 1.7 / 16.6 & ~16.2 / 64.3 & ~70.5 / 97.5 & ~91.0 / 99.6 \\
\ru &      CycleGAN    & ~56.6 / 87.1 & ~17.3 / 50.4 & ~36.5 / 61.6 &  ~ 1.7 / 16.7 & ~51.1 / 91.1 & ~57.6 / 90.7 & ~88.6 / 98.7 \\
\ru &       StarGAN    & ~ 0.8 / 27.2 & ~68.3 / 98.2 & ~99.8 /100.0 &  ~ 1.7 / 16.8 & ~65.5 / 94.6 & ~81.2 / 99.4 & ~97.9 /100.0 \\
\ru &        RelGAN    & ~87.0 / 99.6 & ~43.5 / 92.7 & ~99.9 /100.0 &  ~ 1.7 / 16.8 & ~24.9 / 71.7 & ~52.1 / 95.7 & ~93.7 / 99.8 \\
\ru &        GauGAN    & ~18.9 / 61.9 & ~ 1.2 / 17.0 & ~ 1.5 / 30.9 &  ~ 1.7 / 16.8 & ~61.3 / 93.8 & ~77.4 / 97.5 & ~89.6 / 99.5 \\ \hline \hline
\ru \multirow{3}{*}{ \rotatebox{90}{High R.~}}
    & ProGAN           & ~83.2 / 88.0 & ~29.2 / 92.2 & 100.0 /100.0 &  ~ 0.9 /  9.0 & 100.0 /100.0 & 100.0 /100.0 & 100.0 /100.0  \\
\ru & StyleGAN         & ~88.5 / 96.9 & ~16.6 / 64.0 & ~51.3 / 85.3 &  ~ 2.2 / 21.7 & ~84.5 / 98.9 & ~94.5 / 99.8 & ~99.0 /100.0 \\ 
\ru & StyleGAN2        & ~98.1 / 99.8 & ~ 3.4 / 54.3 & ~15.9 / 69.6 &  ~ 0.1 /  1.4 & ~50.2 / 87.6 & ~97.1 / 99.8 & ~93.2 / 99.7  \\ \hline \hline
\ru & avg.             & ~51.1 / 71.2 & ~25.1 / 61.0 & ~50.3 / 80.0 &  ~ 1.3 / 13.0 & ~58.5 / 88.1 & ~79.6 / 97.5 & ~{\bf 93.4 / 99.6} \\
\hline 
    \end{tabular}
    \vspace{2mm}
    \caption{Pd@1\% and Pd@10\% for all the methods that have been trained only on ProGAN dataset.}
    \label{tab:my_label_1}
\end{table*}

\section{Experimental results}
In this Section we report the experimental results of our method and show comparisons with state-of-the-art. 
For a fair comparison, we trained both our proposal and the competitors on the same single dataset. 
This allows to easily test the generalization ability of the CNN models. 
More precisely, we use the training set provided in \cite{Wang2020} 
comprising 362K real images extracted from various object categories of the LSUN dataset and 362K images generated by 20 ProGAN~\cite{karras2018progressive} models.
All images have the same resolution of 256$\times$256 pixels.

To test generalization, we adopt the same setting considered in~\cite{gragnaniello2021gan} 
and include both low resolution (256$\times$256) and high resolution (1024$\times$1024) images 
coming from GAN architectures never seen in training:
StyleGAN~\cite{karras2019style},
StyleGAN2~\cite{karras2020analyzing},
BigGAN~\cite{brock2018large},
CycleGAN~\cite{zhu2017unpaired},
StarGAN~\cite{choi2018stargan},
RelGAN~\cite{wu2019relgan}, and
GauGAN~\cite{park2019SPADE}.
Overall, we have a total of 39K synthetic images.
For the pristine data we consider different sets based on the image resolution:
ImageNet, COCO~\cite{lin2014microsoft}, and Unpaired real dataset~\cite{zhu2017unpaired} for low-resolution images and  the RAISE dataset~\cite{dang2015raise} for high resolution ones.
To avoid any form of polarization, we avoid using the same source of real images (e.g. LSUN) for training and testing. 
Overall we have a total of 11.1K low-resolution and 7.8K high-resolution real images.

For comparisons we consider the following state-of-the-art GAN detectors:
Spec \cite{Zhang2019}, 
FFD \cite{dang2020detection},
PatchForensics \cite{chai2020makes},
CNN-DCT \cite{frank2020leveraging},
Wang2020 \cite{Wang2020}, 
no-down \cite{gragnaniello2021gan}.
Note that for PatchForensics we use the original weights of the proposal, also obtained by training the network on ProGAN.

\begin{figure}[t!]
    \centering
    \includegraphics[width=0.95\linewidth,trim=0 10 0 0,clip]{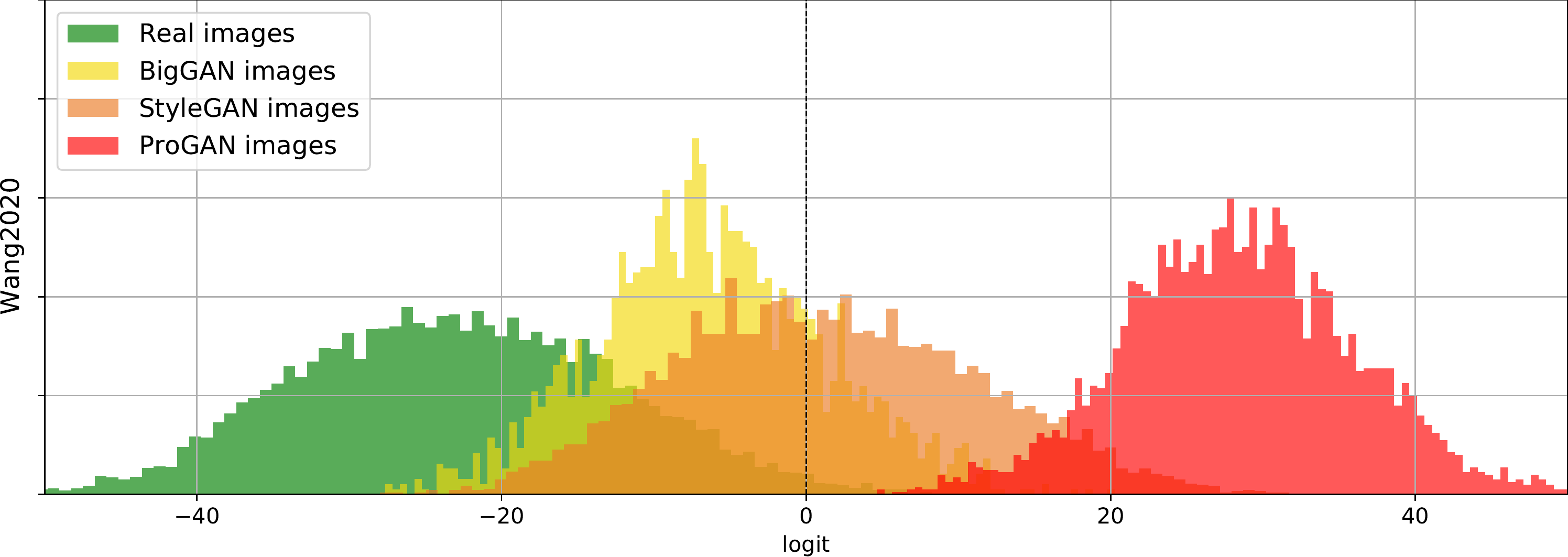} \\
    \includegraphics[width=0.95\linewidth]{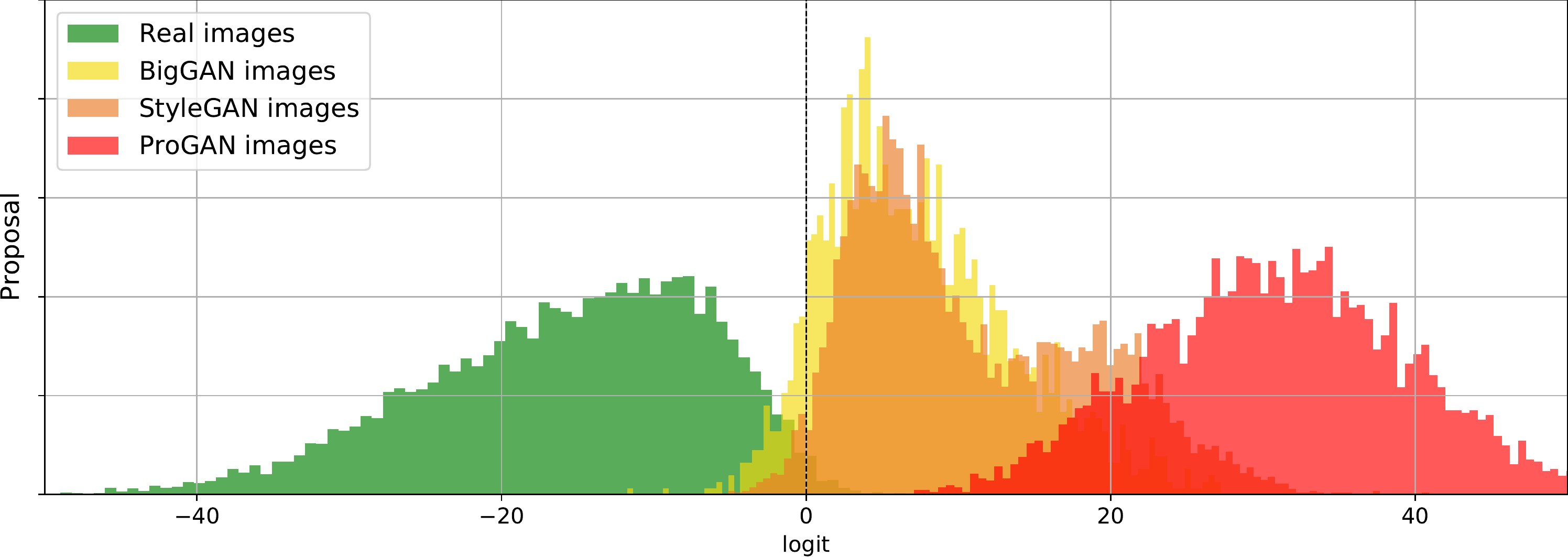}
    \caption{Distribution of network output. For Wang2020 (top) a different threshold is necessary for each GAN while for the proposed method (bottom) a single threshold works well in all cases.}
    \label{fig:hist}
\end{figure}

Our first set of experiments aims at testing generalization on GAN architectures never seen in training.
In Tab.\ref{tab:AUC-Acc} we report both AUC and Accuracy.
Except for CNN-DCT and FFD, all methods ensure very high AUC values, going from 0.888 to 0.997, on the average.
However, such large AUC values often correspond to very low accuracies (e.g., PatchForensics on StyleGAN).
The point is that a large AUC means only that a good threshold exists which separates well synthetic from real images.
Note that this threshold is unknown and may vary from case to case.
Lacking this prior information, we used a fixed 0.5 threshold in the experiment, obtaining several disappointing results.
The problem is well highlighted by the histograms of scores reported in Fig.\ref{fig:hist}.
For Wang2020 \cite{Wang2020}, real and fake images cannot be accurately separated with a single threshold for all cases.
On the contrary, for our proposal, a fixed threshold separates well real data from synthetic images coming from different architectures.
Accordingly, in Tab.\ref{tab:AUC-Acc} the proposed method achieves both high AUC and high accuracy, showing a good generalization ability.

To gain further insights, we also show in Tab.\ref{tab:my_label_1} the probability of detection obtained by setting the false alarm rate at a fixed value of 10\% (Pd@10\%) and 1\% (Pd@1\%).
This is a very appropriate performance measure for real-world operations, where a huge number of images, mostly real, must be analyzed, and a low false alarm rate is mandatory to avoid being swamped by false positives.
Note that the threshold implicitly defined by this metric depends solely on the real images.
Overall performance drops for most of the compared methods. 
Acceptable values are obtained only with PatchForensics and Wang2020, with an average Pd@10\% of 80.0 and 88.1, respectively. 
On the contrary, a very good performance is obtain with no-down (97.5) and our proposal (99.6) which share the choice of avoiding sub-sampling in the first layers.
For the proposed method, even Pd@1\% exceeds 90, on the average, outperforming all reference methods.

\begin{figure}[h!]
    \centering
    \includegraphics[width=0.95\linewidth,trim=0 0 0 0, clip]{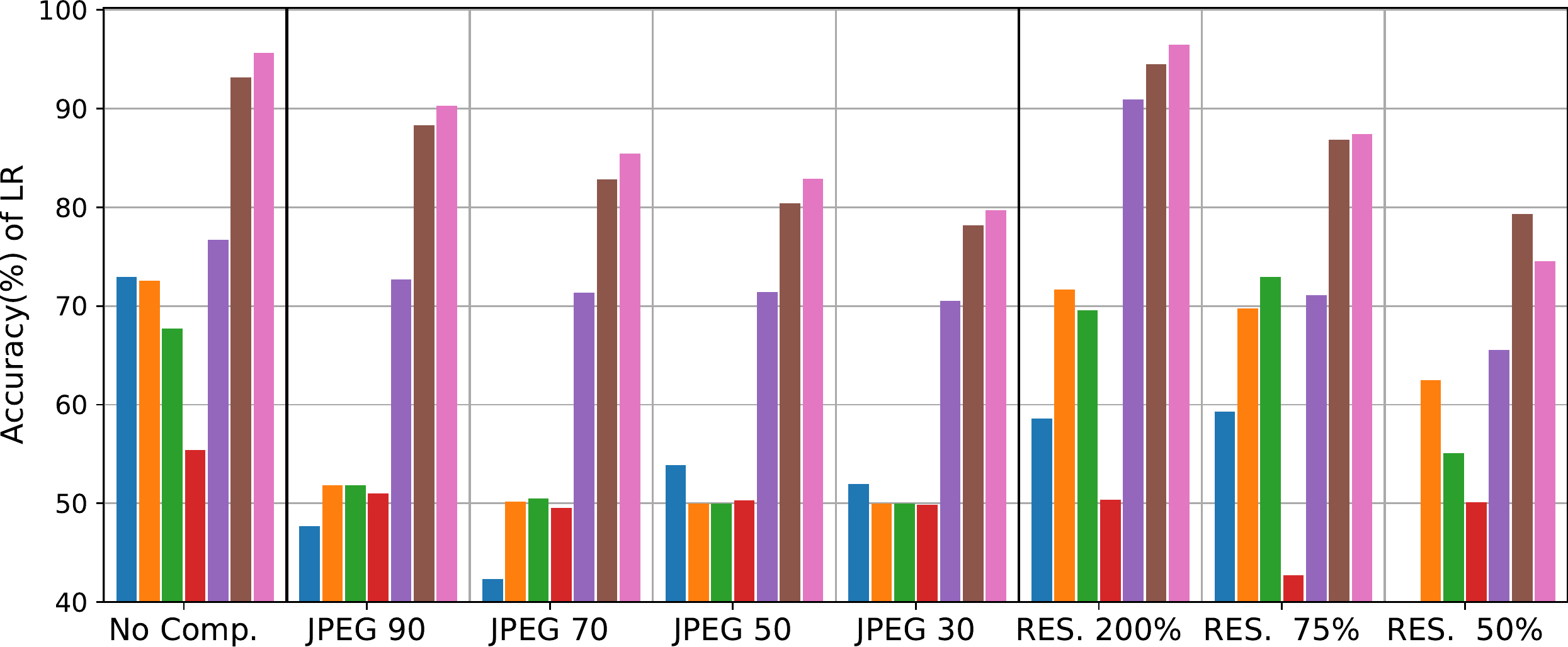} \\[2mm]
    \includegraphics[width=0.95\linewidth,trim=0 0 0 0, clip]{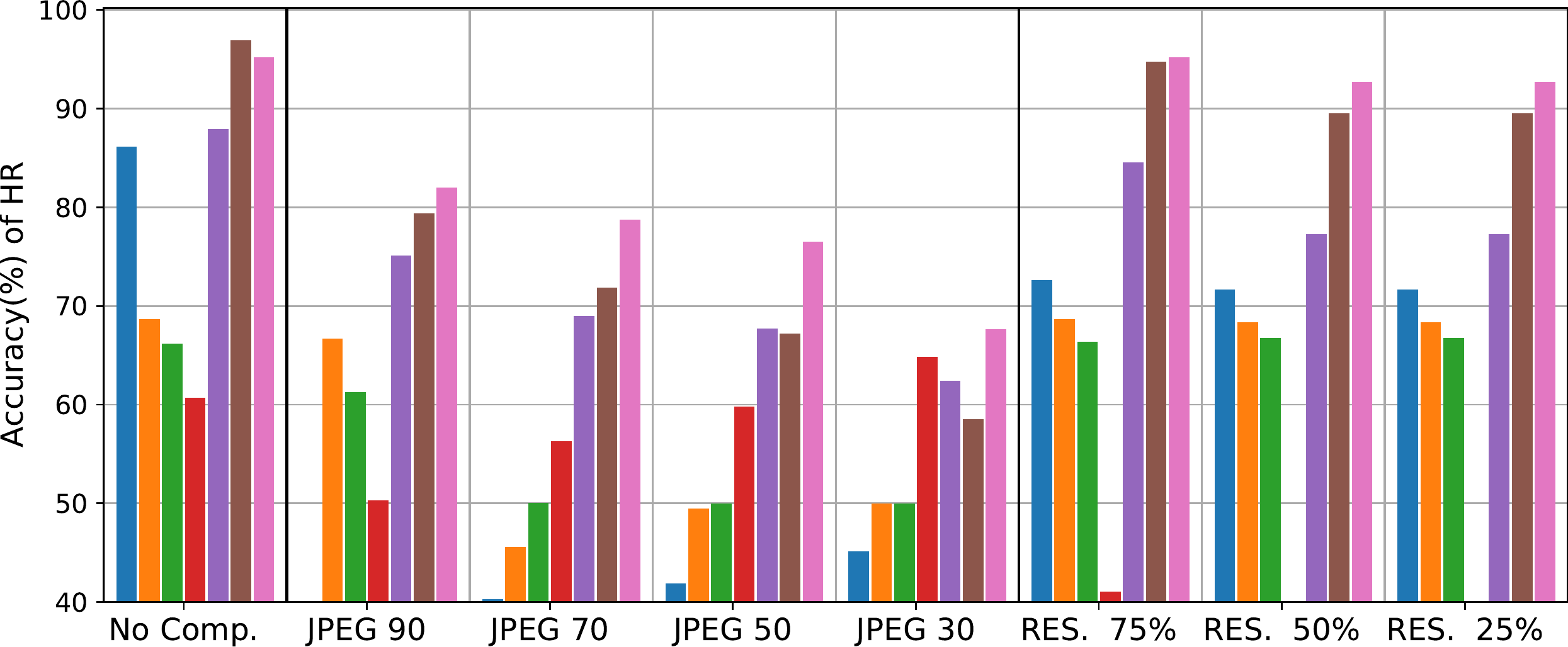} \\[1mm]
    \includegraphics[width=1.1\linewidth,trim=30 0 0 0, clip]{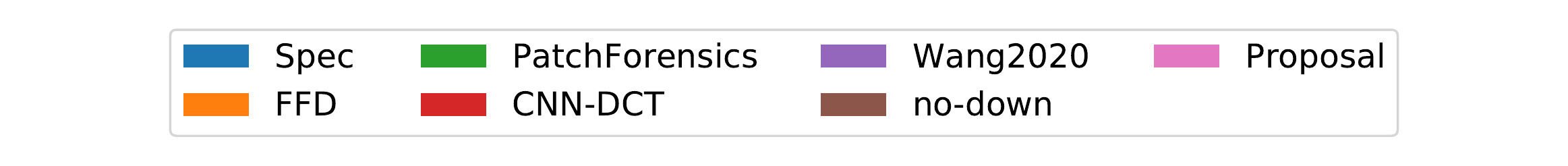} \\
    \caption{Performance in terms of accuracy on low-resolution (LR) images (top) and high resolution (HR) images (bottom)
    by varying the JPEG compression level and rescaling the image of different factors.}
    \label{fig:my_label}
\end{figure}

Finally, to test robustness to typical impairments occurring on social networks, we applied post-processing operations like JPEG compression (with different quality factors) and image resizing (at different scales).
Results in terms of accuracy are compared in Fig.\ref{fig:my_label}.
For several methods, performance reduces dramatically in the presence of compression and resizing,
and remains almost stable only for methods that use strong augmentation and overall for our solution that achieves on the average the best performance.

\section{Conclusion}
We proposed a new method based on contrastive learning for GAN image detection.
It generalizes well to GAN architectures never seen in training and is robust to disruptive forms of post-processing, like compression and resizing. Moreover, it represents a first step towards a universal detector. Among the many problems that remain to investigate, beyond mere performance, we want to address robustness to malicious operations, like adversarial attacks.

\section{Acknowledgement}
This material is based on research sponsored by the Defense Advanced Research Projects Agency (DARPA) and the Air Force Research Laboratory (AFRL) under agreement number FA8750-20-2-1004.
The views and conclusions contained herein are those of the authors and should not be interpreted as necessarily representing the official policies or endorsements, either expressed or implied, of DARPA and AFRL or the U.S. Government. In addition, this work is supported by Google and by the PREMIER project, funded by the Italian Ministry of Education, University, and Research within the PRIN 2017 program.

\clearpage
\bibliographystyle{IEEEtran}
\bibliography{IEEEabrv,ref}

\end{document}